\documentclass[11pt]{article}
\newif\ifcifar
\cifartrue 
\usepackage{booktabs} 
\usepackage{nnonecolumn}
\usepackage{xcolor}

\usepackage[utf8]{inputenc} 
\usepackage[T1]{fontenc}    
\usepackage{booktabs}       
\usepackage{amsfonts}       
\usepackage{nicefrac}       
\usepackage{microtype}      
\usepackage{amsthm}
\usepackage{amssymb}
\usepackage{amsmath}
\usepackage{bm}
\usepackage{mathtools}
\usepackage{enumitem}
\usepackage{wrapfig}
\usepackage{natbib}[sort&compress]
\setcitestyle{authoryear,open={(},close={)}}

\DeclarePairedDelimiterX{\infdivx}[2]{(}{)}{%
  #1\;\delimsize\|\;#2%
}

\DeclarePairedDelimiter{\norm}{\lVert}{\rVert}

\newcommand{\Pdata}{\mathbb{P}_{\text{data}}}
\newcommand{\Dx}{D(\bm{x})}
\newcommand{\DGz}{D(G(\bm{z}))}
\newcommand{\Dxc}{D(\bm{x}\oplus\bm{c})}
\newcommand{\Gzc}{G(\bm{z}\oplus\bm{c})}
\newcommand{\DGzc}{D(\Gzc)}

\renewcommand{\vec}[1]{\mathbf{#1}}

\begin{document}
\title{Evolving GAN Formulations for Higher Quality Image Synthesis}
\author{Santiago Gonzalez$^{1,2,}$\footnote{Current affiliation: Apple Inc.}~, Mohak Kant$^2$, Risto Miikkulainen$^{1,2}$\\
$^1$The University of Texas at Austin, $^2$Cognizant AI Labs}

\maketitle

\begin{abstract}
Generative Adversarial Networks (GANs) have extended deep learning to complex generation and translation tasks across different data modalities. However, GANs are notoriously difficult to train: Mode collapse and other instabilities in the training process often degrade the quality of the generated results, such as images. This paper presents a new technique called TaylorGAN for improving GANs by discovering customized loss functions for each of its two networks. The loss functions are parameterized as Taylor expansions and optimized through multiobjective evolution. On an image-to-image translation benchmark task, this approach qualitatively improves generated image quality and quantitatively improves two independent GAN performance metrics. It therefor forms a promising approach for applying GANs to more challenging tasks in the future.
\end{abstract}

\ifcifar
%
%
\fi

\section{Introduction}

Generative Adversarial Networks (GANs) have recently emerged as a promising technique for building models that generate new samples according to a distribution within a dataset. In GANs, two separate networks---a generator and a discriminator---are trained in tandem in an adversarial fashion: The generator attempts to synthesize samples that the discriminator believes is real, while the discriminator attempts to differentiate between samples from the generator and samples from a ground-truth dataset. However, GANs are challenging to train. Training often suffers from instabilities that can lead to low-quality and potentially low-variety generated samples. These difficulties have lead many researchers to try formulating better GANs, primarily by designing new generator and discriminator loss functions by hand.

Neuroevolution may potentially offer a solution to this problem. It has recently been extended from optimizing network weights and topologies to designing deep learning architectures \citep{stanley:naturemi19,real:aaai19,liang:gecco19}.
Advances in this field,---known as evolutionary metalearning---have resulted in designs that outperform those that are manually-tuned. One particular family of techniques---loss-function metalearning---has allowed for neural networks to be trained more quickly, with higher accuracy, and better robustness \citep{gonzalez2019glo,taylorglo}. Perhaps loss-function metalearning can be adapted to improve GANs?

In this paper, such a technique is developed to evolve entirely new GAN formulations that outperform the standard Wasserstein loss. Leveraging the TaylorGLO loss-function parameterization approach \citep{taylorglo}, separate loss functions are constructed for the two GAN networks. A genetic algorithm is then used to optimize their parameters against two non-differentiable objectives. A composite transformation of these objectives \citep{shahrzad2018enhanced} is further used to enhance the multiobjective search.

This TaylorGAN approach is evaluated experimentally in an image-to-image translation benchmark task where the goal is to generate photorealistic building images based on a building segment map. The CMP Facade dataset \citep{facades} is used as the training data and the pix2pix-HD conditional GAN \citep{pix2pixhd} as the generative model.  The approach is found to both qualitatively enhance generated image quality and quantitatively improve the two metrics.
The evaluation thus demonstrates how evolution can improve a leading conditional GAN design by replacing manually designed loss functions with those optimized by a multiobjective genetic algorithm.

Section~\ref{sec:background} reviews key literature in GANs, motivating the evolution of their loss functions. The next section describes the TaylorGLO metalearning technique for optimizing loss-functions in general. Section~\ref{sec:technique} introduces the TaylorGAN variation of it, focusing on how the TaylorGLO loss-function parameterization is leveraged for evolving GANs. Section~\ref{sec:expsetup} details the experimental configuration and evaluation methodologies. In Section~\ref{sec:results}, TaylorGAN's efficacy is evaluated on the benchmark task. Section~\ref{sec:discussion} places these these findings in the general context of the GAN literature and describes potential avenues for future work.

\section{Variations of GAN Architectures}
\label{sec:background}

Generative Adversarial Networks \citep[GANs;][]{goodfellow2014gan}, are a type of generative model consisting of a pair of networks, a generator and discriminator, that are trained in tandem. GANs are a modern successor to Variational Autoencoders \citep[VAEs;][]{kingma2013vae} and Boltzmann Machines \citep{hinton1983boltzmann}, including Restricted Boltzmann Machines \citep{smolensky1986rbm} and Deep Boltzmann Machines \citep{salakhutdinov2009deepboltzmann}.

The following subsections review prominent GAN methods. Key GAN formulations, and the relationships between them, are described. Consistent notation (shown in Table~\ref{tab:gan_notation}) is used, consolidating the extensive variety of notation in the field. 

\begin{table}[ht]
\caption{GAN Notation Decoder}
\vspace*{-2ex}
\begin{center}
\begin{tabular}{cl}
\toprule
\textbf{Symbol} & \textbf{Description}\\
\midrule
    $G(\bm{x},\theta_G)$ & Generator function \\
    $D(\bm{z},\theta_D)$ & Discriminator function \\
    $\Pdata$ & Probability distribution of the original data \\
    $\mathbb{P}_z$ & Latent vector noise distribution \\
    $\mathbb{P}_g$ & Probability distribution of $G(\bm{z})$ \\
    $\bm{x}$ & Data, where $\bm{x}\sim \Pdata$ \\
    $\bm{\tilde{x}}$ & Generated data \\
    $\bm{z}$ & Latent vector, where $\bm{z}\sim \mathbb{P}_z$  \\
    $\bm{c}$ & Condition vector \\
    $\lambda$ & Various types of weights / hyperparameters\\
\bottomrule
\end{tabular}
\end{center}
\label{tab:gan_notation}
\end{table}

\subsection{Overview}


A GAN's generator and discriminator are set to compete with each other in a minimax game, attempting to reach a Nash equilibrium \citep{nash1951non,heusel2017gans}. Throughout the training process, the generator aims to transform samples from a prior noise distribution into data, such as an image, that tricks the discriminator into thinking it has been sampled from the real data's distribution. Simultaneously, the discriminator aims to determine whether a given sample came from the real data's distribution, or was generated from noise.


Unfortunately, GANs are difficult to train, frequently exhibiting instability, i.e., mode collapse, where all modes of the target data distribution are not fully represented by the generator \citep{radford2015dcgan,metz2016unrolledgan,isola2016imagetoimage,mao2017lsgan,arjovsky2017wgan,gulrajani2017improvedwgan,mao2018lsganeffectiveness}. GANs that operate on image data often suffer from visual artifacts and blurring of generated images \citep{isola2016imagetoimage,odena2016deconvolution}. Additionally, datasets with low variability have been found to degrade GAN performance \citep{mao2018lsganeffectiveness}.

GANs are also difficult to evaluate quantitatively, typically relying on metrics that attempt to embody vague notions of quality. Popular GAN image scoring metrics, for example, have been found to have many pitfalls, including cases where two samples of clearly disparate quality may have similar values \citep{borji2019proscons}.

\subsection{Original Minimax and Non-Saturating GAN}

Using the notation described in Table~\ref{tab:gan_notation}, the original minimax GAN formulation by \cite{goodfellow2014gan} can be defined as
\begin{equation}
\min_{\theta_G}\max_{\theta_D}\; \mathbb{E}_{\bm{x} \sim \Pdata}\left[\log \Dx\right] + \mathbb{E}_{\bm{z} \sim \mathbb{P}_z}\left[\log \left(1- \DGz\right)\right]    \;.
\end{equation}
This formulation can be broken down into two separate loss functions, one each for the discriminator and generator:
\begin{eqnarray}
\mathcal{L}_D &=& -\frac{1}{n}\sum^n_{i=1} \left[  \log D(x_i) + \log(1-D(G(z_i)))  \right]   \;\text{, and} \\
\mathcal{L}_G &=& \frac{1}{n}\sum^n_{i=1} \log(1-D(G(z_i)))    \;.
\end{eqnarray}
The discriminator's loss function is equivalent to a sigmoid cross-entropy loss when thought of as a binary classifier. \cite{goodfellow2014gan} proved that training a GAN with this formulation is equivalent to minimizing the Jensen-Shannon divergence between $\mathbb{P}_g$ and $\Pdata$, i.e. a symmetric divergence metric based on the Kullback-Leibler divergence.

In the above formulation the generator's loss saturates quickly since the discriminator learns to reject the novice generator's samples early on in training. To resolve this problem, Goodfellow et al.\ provided a second ``non-saturating'' formulation with the same fixed-point dynamics, but better, more intense gradients for the generator early on:
\begin{eqnarray}
&\max_{\theta_D}& \mathbb{E}_{\bm{x} \sim \Pdata}\left[\log \Dx\right] + \mathbb{E}_{\bm{z} \sim \mathbb{P}_z}\left[\log \left(1- \DGz\right)\right]  \;, \\
&\max_{\theta_G}& \mathbb{E}_{\bm{z} \sim \mathbb{P}_z}\left[\log \DGz\right]   \;.
\end{eqnarray}

Each GAN training step consists of training the discriminator for $k$ steps, while sequentially training the generator for only one step. This difference in steps for both networks helps prevent the discriminator from learning too quickly and overpowering the generator.

Alternatively, Unrolled GANs \citep{metz2016unrolledgan} aimed to prevent the discriminator from overpowering the generator by using a discriminator which has been unrolled for a certain number of steps in the generator's loss, thus allowing the generator to train against a more optimal discriminator. More recent GAN work instead uses a two time-scale update rule \cite[TTUR;][]{heusel2017gans}, where the two networks are trained under different learning rates for one step each. This approach has proven to converge more reliably to more desirable solutions.

Unfortunately, with both minimax and non-saturating GANs the generator gradients vanish for samples that are on the correct side of the decision boundary but far from the true data distribution \citep{mao2017lsgan,mao2018lsganeffectiveness}. The Wasserstein GAN, described next, is designed to solve this problem.

\subsection{Wasserstein GAN}

The Wasserstein GAN \citep[WGAN;][]{arjovsky2017wgan} is arguably one of the most impactful developments in the GAN literature since the original formulation by \cite{goodfellow2014gan}. WGANs minimize the Wasserstein-1 distance between $\mathbb{P}_g$ and $\Pdata$, rather than the Jensen-Shannon divergence, in an attempt to avoid vanishing gradient and mode collapse issues. In the context of GANs, the Wasserstein-1 distance can be defined as
\begin{equation}
W(\mathbb{P}_g,\Pdata) = \inf_{\gamma \in \Pi(\mathbb{P}_g,\Pdata)} \mathbb{E}_{(\bm{u},\bm{v})\sim \gamma} \left[\norm{\bm{u}-\bm{v}}  \right]   \;,
\end{equation}
where, $\gamma(\bm{u},\bm{v})$ represents the amount of mass that needs to move from $\bm{u}$ to $\bm{v}$ for $\mathbb{P}_g$ to become $\Pdata$. This formulation with the infimum is intractable, but the Kantorovich-Rubinstein duality \citep{villani2009wasserstein} with a supremum makes the Wasserstein-1 distance tractable, while imposing a 1-Lipschitz smoothness constraint:
\begin{equation}
W(\mathbb{P}_g,\Pdata) = \sup_{\norm{f}_L \leq 1} \mathbb{E}_{\bm{u}\sim \mathbb{P}_g} \left[f(\bm{u})  \right] - \mathbb{E}_{\bm{u}\sim \Pdata} \left[f(\bm{u})  \right]  \;,
\end{equation}
which translates to the training objective
\begin{equation}
\min_{\theta_G}\max_{\theta_D \in \Theta_D} \mathbb{E}_{\bm{x} \sim \Pdata}\left[ \Dx\right] - \mathbb{E}_{\bm{z} \sim \mathbb{P}_z}\left[\DGz\right]  \; ,
\end{equation}
where $\Theta_D$ is the set of all parameters for which $D$ is a 1-Lipschitz function.

WGANs are an excellent example of how generator and discriminator loss functions can profoundly impact the quality of generated samples and the prevalence of mode collapse. However, the WGAN has a 1-Lipschitz constraint that needs to be maintained throughout training for the formulation to work. WGANs enforce the constraint via gradient clipping, at the cost of requiring an optimizer that does not use momentum, i.e., RMSProp \citep{rmsprop} rather than Adam \citep{adam}.

To resolve the issues caused by gradient clipping, a subsequent formulation, WGAN-GP \citep{gulrajani2017improvedwgan}, added a gradient penalty regularization term to the discriminator loss:
\begin{equation}
GP = \lambda\; \mathbb{E}_{\bm{\hat{x}} \sim \mathbb{P}_{\hat{x}}} \left[\left( \norm{\nabla_{\bm{\hat{x}}} D(\bm{\hat{x}})}_2 - 1 \right)^2\right]   \;,
\end{equation}
where $\mathbb{P}_{\hat{x}}$ samples uniformly along lines between $\Pdata$ and $\mathbb{P}_g$. The gradient penalty enforces a soft Lipschitz smoothness constraint, leading to a more stationary loss surface than when gradient clipping is used, which in turn makes it possible to use momentum-based optimizers. The gradient penalty term has even been successfully used in non-Wasserstein GANs \citep{fedus2018manypaths,mao2018lsganeffectiveness}. However, gradient penalties can increase memory and compute costs \citep{mao2018lsganeffectiveness}.

\subsection{Least-Squares GAN}

Another attempt to solve the issue of vanishing gradients is the Least-Squares GAN \citep[LSGAN][]{mao2017lsgan}. It defines the training objective as
\begin{eqnarray}
&\min_{\theta_D}& \frac{1}{2}\; \mathbb{E}_{\bm{x} \sim \Pdata} \left[\left(\Dx-b\right)^2\right] + \mathbb{E}_{\bm{z} \sim \mathbb{P}_z} \left[\left(\DGz-a\right)^2\right]    \;, \\
&\min_{\theta_G}& \frac{1}{2}\; \mathbb{E}_{\bm{z} \sim \mathbb{P}_z} \left[\left(\DGz-c\right)^2\right]  \;,
\end{eqnarray}
where $a$ is the label for generated data, $b$ is the label for real data, and $c$ is the label that $G$ wants to trick $D$ into believing for generated data. In practice, typically $a = 0, b = 1, c = 1$. However, subsequently, $a=-1,b=-1,c=0$ were found to result in faster convergence, making it the recommended parameter setting \citep{gulrajani2017improvedwgan}. Training an LSGAN was shown to be equivalent to minimizing the Pearson $\chi^2$ divergence \citep{pearson1900x} between $\Pdata + \mathbb{P}_g$ and $2*\mathbb{P}_g$. Generated data quality can oscillate throughout the training process \citep{mao2018lsganeffectiveness}, indicating a disparity between data quality and loss.

\subsection{Conditional GAN}

Traditional GANs learn how to generate data from a latent space, i.e. an embedded representation of the training data that the generator constructs. Typically, the elements of a latent space have no immediately intuitive meaning \citep{chen2016infogan,larsen2015ganvae}. Thus, GANs can generate novel data, but there is no way to steer the generation process to generate particular types of data. For example, a GAN that generates images of human faces cannot be explicitly told to generate a face with a particular hair color or of a specific gender. While techniques have been developed to analyze this latent space \citep{volz2018evolvingmario,li2019gagan}, or build more interpretable latent spaces during the training process \citep{chen2016infogan}, they do not necessarily translate a human's prior intuition correctly or make use of labels when they are available. To tackle this problem, Conditional GANs, first proposed as future work by \cite{goodfellow2014gan} and subsequently developed \cite{mirza2014conditional}, allow directly targetable features (i.e., conditions) to be an integral part of the generator's input.

The conditioned training objective for a minimax GAN can be defined, without loss of generality, as
\begin{equation}
\min_{\theta_G}\max_{\theta_D}\; \mathbb{E}_{\bm{x} \sim \Pdata}\left[\log \Dxc\right] + \mathbb{E}_{\bm{z} \sim \mathbb{P}_z}\left[\log \left(1- \DGzc\right)\right] \;,
\end{equation}
where $\bm{z}\oplus\bm{c}$ is basic concatenation of vectors. During training, the condition vector, $\bm{c}$, arises from the sampling process that produces each $\bm{x}$. This same framework can be used to design conditioned variants of other GAN formulations. 

Conditional GANs have enjoyed great successes as a result of their flexibility, even in the face of large, complex condition vectors, which may even be whole images. They enable new applications for GANs, including repairing software vulnerabilities \cite[framed as sequence to sequence translation;][]{harer2018softwaregan}, integrated circuit mask design \citep{alawieh2019gansraf}, and image to image translation \citep{isola2016imagetoimage}---the generation of images given text \citep{reed2016cgantext}---which is used as the target setting for this paper. Notably, conditional GANs can increase the quality of generated samples for labeled datasets, even when conditioned generation is not needed \citep{vandenoord2016pixelcnn}. Conditional GANs are therefore used as the platform for the TaylorGAN technique described in the next section.

\subsection{Opportunity: Optimizing Loss Functions}

The GAN formulations described above all have one property in common: The generator and discriminator loss functions have been arduously derived by hand. A GAN's performance and stability is greatly impacted by the choice of loss functions. Different regularization terms, such as the aforementioned gradient penalty can also affect a GAN's training. These elements of the GAN are typically designed to minimize a specific divergence. However, a GAN does not need to decrease a divergence at every step in order to reach the Nash equilibrium \citep{fedus2018manypaths}. In this situation, an automatic loss-function optimization system may find novel GAN loss functions with more desirable properties. Such a system is presented in Section~\ref{sec:technique} and evaluated on conditional GANs in Section~\ref{sec:results}. 
The basic method for evolving loss functions, TaylorGLO, is reviewed in the next section.

\section{Evolution of Loss Functions}
\label{sec:taylorglo_method}

Loss-function metalearning makes it possible to regularize networks automatically; TaylorGLO is a flexible and scalable implementation of this idea based on multivariate Taylor expansions.

\begin{figure}
  \centering
  \includegraphics[width=0.5\linewidth]{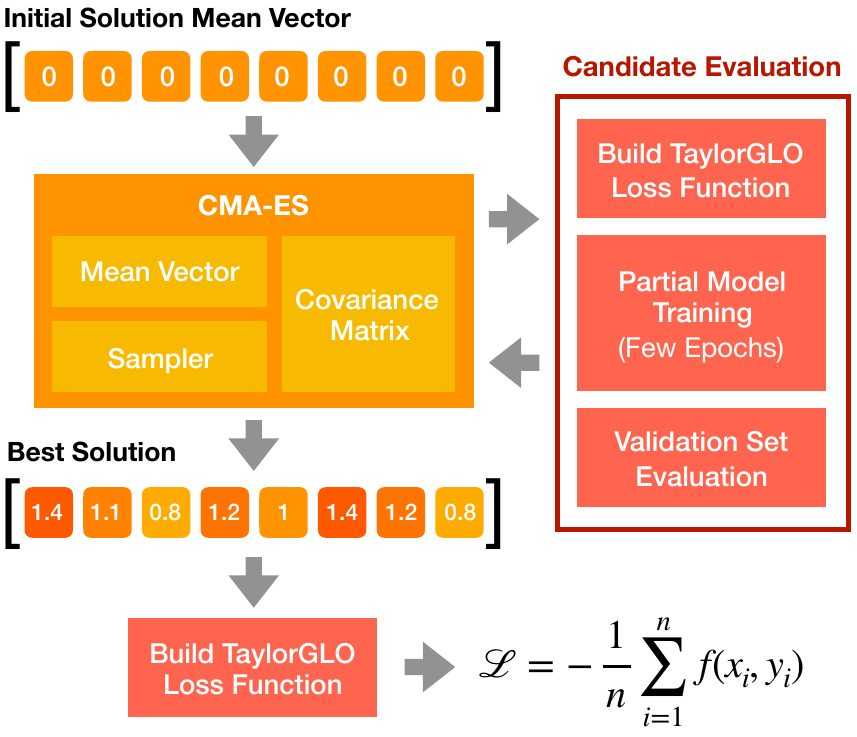}
  \caption{The TaylorGLO method \citep{taylorglo}. Loss functions are represented by fixed-size vectors whose elements parameterize modified Taylor polynomials. Starting with a population of initially unbiased loss functions (i.e., vectors around the origin), CMA-ES optimizes their Taylor expansion parameters in order to maximize validation accuracy after partial training. The candidate with the highest accuracy is chosen as the final, best solution. This approach biases the search towards functions with useful properties, and is also amenable to theoretical analysis, as shown in this paper.}
  \label{fig:overview}
\end{figure}

\subsection{Motivation}

Loss-function metalearning for deep networks was first introduced by \citet{gonzalez2019glo} as an automatic way to find customized loss functions that optimize a performance metric for a model. The technique, a genetic programming approach named GLO, discovered one particular loss function, Baikal, that improves classification accuracy, training speed, and data utilization. Intuitively, Baikal achieved these properties through a form of regularization that ensured the model would not become overly confident in its predictions. That is, instead of monotonically decreasing the loss when the output gets closer to the correct value, Baikal loss increases rapidly when the output is almost correct, thus discouraging extreme accuracy.

TaylorGLO \citep{taylorglo} is a scalable reformulation of the GLO approach.\footnote{Open-source code for TaylorGLO is available at \texttt{https://github.com/cognizant-ai-labs/taylorglo}.}  Instead of trees evolved through genetic programming, TaylorGLO represents loss functions as parameterizations of multivariate Taylor polynomials. It is then possible to evolve the parameters directly with CMA-ES, which makes it possible to scale to models with millions of trainable parameters and a variety of deep learning architectures.

\subsection{Loss Functions as Multivariate Taylor Expansions}

\newcommand{\Tx}[1][-\theta_0]{(x_i #1)}
\newcommand{\Ty}[1][-\theta_1]{(y_i #1)}

Taylor expansions \citep{taylor1715methodus} represent differentiable functions within the neighborhood of a point using a polynomial series. In the univariate case, given a $C^{k_{\text{max}}}$ smooth (i.e., first through $k_{\text{max}}$ derivatives are continuous), real-valued function, $f(x): \mathbb{R}\to\mathbb{R}$, a $k$th-order Taylor approximation at point $a\in \mathbb{R}$, $\hat{f}_k(x,a)$, where $0\leq k \leq k_{\text{max}}$, can be constructed as
\begin{equation}
\hat{f}_k(x,a) = \sum_{n=0}^k \frac{1}{n!} f^{(n)}(a) (x-a)^n    .
\end{equation}
This formulation can be extended to the multivariate case by defining an $n$th-degree multi-index, $\alpha = (\alpha_1,\alpha_2,\ldots,\alpha_n)$, where $\alpha_i \in \mathbb{N}_0$, $|\alpha| = \sum_{i=1}^n \alpha_i$, $\alpha! = \prod_{i=1}^n \alpha_i!$. $\vec{x}^\alpha = \prod_{i=1}^n x_i^{\alpha_i}$, and $\vec{x} \in \mathbb{R}^n$. Multivariate partial derivatives can be concisely written using a multi-index as
\begin{equation}
\partial^\alpha f = \partial_1^{\alpha_1}\partial_2^{\alpha_2}\cdots\partial_n^{\alpha_n} f = \frac{\partial^{|\alpha|}}{\partial x_1^{\alpha_1}\partial x_2^{\alpha_2}\cdots\partial x_n^{\alpha_n}}   .
\end{equation}
Thus, discounting the remainder term, the multivariate Taylor expansion for $f(\vec{x})$ at $\vec{a}$ is
\begin{equation}
\hat{f}_k(\vec{x},\vec{a}) = \sum_{\forall \alpha,|\alpha|\leq k} \frac{1}{\alpha!} \partial^{\alpha}f(\vec{a}) (\vec{x}-\vec{a})^\alpha    .
\end{equation}
The unique partial derivatives in $\hat{f}_k$ and $\vec{a}$ are parameters for a $k$th order Taylor expansion. Thus, a $k$th order Taylor expansion of a function in $n$ variables requires $n$ parameters to define the center, $\vec{a}$, and one parameter for each unique multi-index $\alpha$, where $|\alpha|\leq k$. That is: 
\begin{equation}
\#_{\text{parameters}}(n,k) = n + {n+k\choose k} = n + \frac{(n+k)!}{n!\,k!}  \;.
\end{equation}

The multivariate Taylor expansion can be leveraged for loss-function parameterization \citep{taylorglo}. Let an $n$-class classification loss function be defined as $\mathcal{L}_{\text{Log}} = -\frac{1}{n}\sum^{n}_{i=1} f(x_i,y_i)$.
The function $f(x_i,y_i)$ can be replaced by its $k$th-order, bivariate Taylor expansion, $\hat{f}_k(x,y,a_x,a_y)$. For example, a loss function in $\vec{x}$ and $\vec{y}$ has the following third-order parameterization with parameters $\vec{\theta}$ (where $\vec{a} = \left< \theta_0, \theta_1 \right>$):
\begin{equation}
\label{eq:k3example}
\begin{aligned}
\mathcal{L}(\vec{x},\vec{y}) = -\frac{1}{n}\sum^n_{i=1} \Big[      \theta_2 + \theta_3\Ty + \tfrac{1}{2}\theta_4\Ty^2 \\+ \tfrac{1}{6}\theta_5\Ty^3 + \theta_6\Tx 
   + \theta_7\Tx\Ty \\+ \tfrac{1}{2}\theta_8\Tx\Ty^2 + \tfrac{1}{2}\theta_9\Tx^2 \\
   + \tfrac{1}{2}\theta_{10}\Tx^2\Ty + \tfrac{1}{6}\theta_{11}\Tx^3  \Big]
\end{aligned}
\end{equation}

As was shown by \cite{taylorglo}, the technique makes it possible to train neural networks that are more accurate and learn faster than those with tree-based loss function representations. Representing loss functions in this manner guarantees that the functions are smooth, do not have poles, can be implemented through addition and multiplication, and can be trivially differentiated. The search space is locally smooth and has a tunable complexity parameter (the order of expansion), making it possible to find valid loss functions consistently and with high frequency. These properties are not necessarily held by alternative function approximators, such as Fourier expansions, Pad\'{e} approximants, Laurent polynomials, and Polyharmonic splines \citep{taylorglo}.

\subsection{The TaylorGLO Method}

TaylorGLO (Figure~\ref{fig:overview}) aims to find the optimal parameters for a loss function represented as a multivariate Taylor expansion. The parameters for a Taylor approximation (i.e., the center point and partial derivatives) are referred to as $\vec{\theta}_{\hat{f}}$: $\vec{\theta}_{\hat{f}} \in \Theta$, $\Theta = \mathbb{R}^{\#_{\text{parameters}}}$. TaylorGLO strives to find the vector $\vec{\theta}_{\hat{f}}^*$ that parameterizes the optimal loss function for a task. Because the values are continuous, as opposed to discrete graphs of the original GLO, it is possible to use continuous optimization methods.

In particular, Covariance Matrix Adaptation Evolutionary Strategy \citep[CMA-ES;][]{hansen1996cmaes} is a popular population-based, black-box optimization technique for rugged, continuous spaces. CMA-ES functions by maintaining a covariance matrix around a mean point that represents a distribution of solutions. At each generation, CMA-ES adapts the distribution to better fit evaluated objective values from sampled individuals. In this manner, the area in the search space that is being sampled at each step grows, shrinks, and moves dynamically as needed to maximize sampled candidates' fitnesses. TaylorGLO uses the ($\mu/\mu,\lambda$) variant of CMA-ES \citep{hansen2001cmaesmumulambda}, which incorporates weighted rank-$\mu$ updates \citep{hansen2004weightedrankmucmaes} to reduce the number of objective function evaluations needed.

In order to find $\vec{\theta}_{\hat{f}}^*$, at each generation CMA-ES samples points in $\Theta$. Their fitness is determined by training a model with the corresponding loss function and evaluating the model on a validation dataset. Fitness evaluations may be distributed across multiple machines in parallel and retried a limited number of times upon failure. An initial vector of $\vec{\theta}_{\hat{f}} = \vec{0}$ is chosen as a starting point in the search space to avoid bias.

Fully training a model can be prohibitively expensive in many problems. However, performance near the beginning of training is usually correlated with performance at the end of training, and therefore it is enough to train the models only partially to identify the most promising candidates. This type of approximate evaluation is common in metalearning \citep{grefenstette1985genetic, jin2011}. An additional positive effect is that evaluation then favors loss functions that learn more quickly.

For a loss function to be useful, it must have a derivative that depends on the prediction. Therefore, internal terms that do not contribute to $\frac{\partial}{\partial \vec{y}}\mathcal{L}_f(\vec{x},\vec{y})$ can be trimmed away. This step implies that any term $t$ within $f(x_i,y_i)$ with $\frac{\partial}{\partial y_i}t = 0$ can be replaced with $0$. For example, this refinement simplifies Equation~\ref{eq:k3example}, providing a reduction in the number of parameters from twelve to eight:
\begin{equation}
\label{eq:k3taylorglo}
\begin{aligned}
\mathcal{L}(\vec{x},\vec{y}) = -\frac{1}{n}\sum^n_{i=1} \Big[    \theta_2\Ty + \tfrac{1}{2}\theta_3\Ty^2 + \tfrac{1}{6}\theta_4\Ty^3  \\+ \theta_5\Tx\Ty
 + \tfrac{1}{2}\theta_6\Tx\Ty^2 \\+ \tfrac{1}{2}\theta_7\Tx^2\Ty  \Big] \;.
\end{aligned}
\end{equation}

Building on this foundation, the method for evolving GAN formulations is described next.

\section{The TaylorGAN Approach}
\label{sec:technique}

As GANs have grown in popularity, the difficulties involved in training them have become increasingly evident. The loss functions used to train a GAN's generator and discriminator  constitute the core of how GANs are formulated. Thus, optimizing these loss functions jointly can result in better GANs. This section presents an extension of TaylorGLO to evolve loss functions for GANs. Images generated in this way improve both visually and quantitatively, as the experiments in Section~\ref{sec:results} show.

TaylorGLO parameterization represents a loss function as a modified third-degree Taylor polynomial. Such a parameterization has many desirable properties, such as smoothness and continuity, that make it amenable for evolution \citep{taylorglo}.
In TaylorGAN, there are three functions that need to be optimized jointly (using the notation described in Table~\ref{tab:gan_notation}):
\begin{enumerate}
\item The component of the discriminator's loss that is a function of $D(\bm{x})$, the discriminator's output for a real sample from the dataset,
\item The synthetic / fake component of the discriminator's loss that is a function of $D(G(\bm{z}))$, the discriminator's output from the generator that samples $\bm{z}$ from the latent distribution), and
\item The generator's loss, a function of $D(G(\bm{z}))$.
\end{enumerate}
The discriminator's full loss is simply the sum of components (1) and (2). Table~\ref{tab:gan_glo_formulations} shows how existing GAN formulations can be broken down into this tripartite loss.

\begin{table*}[ht]
\caption{\textbf{Interpretation of existing GAN formulations.} These three components are all that is needed to define the discriminator's and generator's loss functions (sans regularization terms). Thus, TaylorGAN can discover and optimize new GAN formulations by jointly evolving three separate functions.}
\begin{center}
\footnotesize
\begin{tabular}{lccc}
\toprule
\textbf{Formulation} & \textbf{Loss $D$ (real)} & \textbf{Loss $D$ (fake)} & \textbf{Loss $G$ (fake)}\\
& $\mathbb{E}_{\bm{x} \sim \Pdata}$ & $\mathbb{E}_{\bm{z} \sim \mathbb{P}_z}$ & $\mathbb{E}_{\bm{z} \sim \mathbb{P}_z}$ \\
\midrule
GAN \citep[minimax;][]{goodfellow2014gan}    & $-\log \Dx$ & $-\log(1-\DGz)$ & $\log(1-\DGz)$ \\
GAN \citep[non-saturating;][]{goodfellow2014gan}   & $-\log \Dx$ & $-\log(1-\DGz)$ & $-\log \DGz$ \\
WGAN \citep{arjovsky2017wgan}  & $-\Dx$  & $\DGz$  & $-\DGz$ \\
LSGAN \citep{mao2017lsgan} & $\frac{1}{2}(\Dx-1)^2$  & $\frac{1}{2}(\DGz)^2$ & $\frac{1}{2}(\DGz-1)^2$ \\
\bottomrule
\end{tabular}
\end{center}
\label{tab:gan_glo_formulations}
\end{table*}

These three functions can be evolved jointly.
GAN loss functions have a single input, i.e. $\Dx$ or $\DGz$. Thus, a set of three third-order TaylorGLO loss functions for GANs requires only 12 parameters to be optimized, making the technique quite efficient.

Fitness for each set of three functions requires a different interpretation than in regular TaylorGLO. Since GANs cannot be thought of as having an accuracy, a different metric needs to be used. The choice of fitness metric depends on the type of problem and target application. In the uncommon case where the training data's sampling distribution is known, the clear choice is the divergence between such a distribution and the distribution of samples from the generator. This approach will be used in the experiments below.

Reliable metrics of visual quality are difficult to define. 
Individual image quality metrics can be exploited by adversarially constructed, lesser-quality images \citep{borji2019proscons}. For this reason, TaylorGAN utilizes a combination of two or more metrics, and multiobjective optimization of them. Good solutions are usually located near the middle of the resulting Pareto front, and they can be found effectively through an objective transformation technique called Composite Objectives \citep{shahrzad2018enhanced}. In this technique, evolution is performed against a weighted sum of metrics. Individual metrics are scaled such that their ranges of typical values match. Thus, if one metric improves, overall fitness will only increases if there is not a comparable regression along another metric.

\section{Experimental Setup}
\label{sec:expsetup}

The technique was integrated into the LEAF evolutionary AutoML framework \citep{liang2019evolutionary}. TaylorGAN parameters were evolved by the LEAF genetic algorithm as if they were hyperparameters. The implementation of CoDeepNEAT \citep{miikkulainen2019evolving} for neural architecture search in LEAF was not used.

The technique was evaluated on the CMP Facade \citep{facades} dataset with a pix2pix-HD model \citep{pix2pixhd}. The dataset consists of only 606 perspective-corrected $256\times 256$ pixel images of building facades. Each image has a corresponding annotation image that segments facades into twelve different components, such as windows and doors. The objective is for the model to take an arbitrary annotation image as an input, and generate a photorealistic facade as output. The dataset was split into a training set with $80\%$ of the images, and validation and testing sets, each with a disjoint $10\%$ of the images.

Two metrics were used to evaluate loss function candidates: (1) structural similarity index measure \citep[SSIM;][]{wang2004ssim} between generated and ground-truth images, and (2) perceptual distance, implemented as the $L_1$ distance between VGG-16 \citep{vgg} ImageNet \citep{ILSVRC15} embeddings for generated and ground-truth images. During evolution, a composite objective \citep{shahrzad2018enhanced} of these two metrics was used to evaluate candidates. The metrics were normalized (i.e., SSIM was multiplied by $17$ and perceptual distance by $-1$) to have a similar impact on evolution.

The target GAN model, pix2pix-HD, is a refinement of the seminal pix2pix model \citep{isola2016imagetoimage}. Both models generate images conditioned upon an input image. Thus, they are trained with paired images. The baseline was trained with the Wasserstein loss \citep{arjovsky2017wgan} and spectral normalization \citep{miyato2018spectral} to enforce the Lipschitz constraint on the discriminator. The pix2pix-HD model is also trained with additive perceptual distance and discriminator feature losses. Both additive losses are multiplied by ten in the baseline. Models were trained for 60 epochs.

When running experiments, each of the twelve TaylorGAN parameters was evolved within $[-10,10]$. The learning rate and weights for both additive losses were also evolved since the baseline values, which are optimal for the Wasserstein loss, may not necessarily be optimal for TaylorGAN loss functions.

\section{Results}
\label{sec:results}

\begin{figure*}
  \centering
  \begin{tabular}{ccccc}

    \multicolumn{5}{l}{ Input:} \\
    \includegraphics[width=0.16\textwidth]{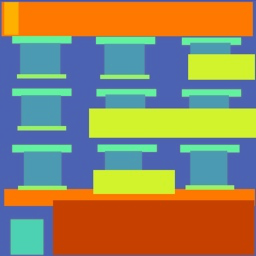} &\
    \includegraphics[width=0.16\textwidth]{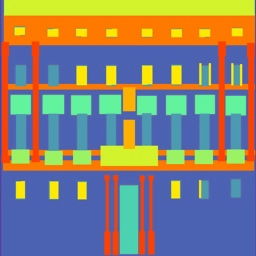} &\
    \includegraphics[width=0.16\textwidth]{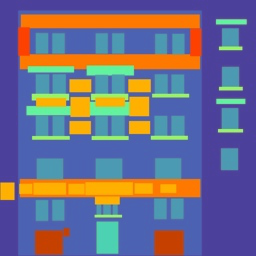} &\
    \includegraphics[width=0.16\textwidth]{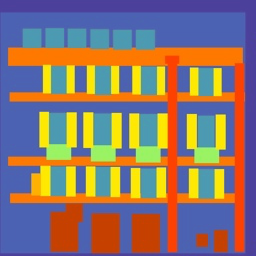} &\
    \includegraphics[width=0.16\textwidth]{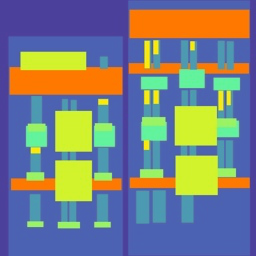} \\

    \multicolumn{5}{l}{ Ground-Truth:} \\
    \includegraphics[width=0.16\textwidth]{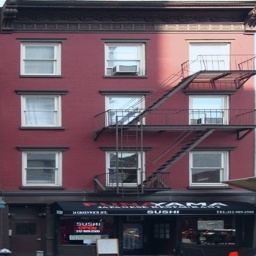} &\
    \includegraphics[width=0.16\textwidth]{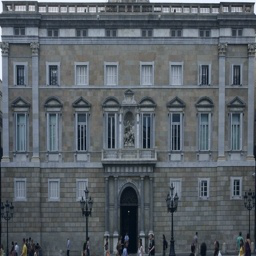} &\
    \includegraphics[width=0.16\textwidth]{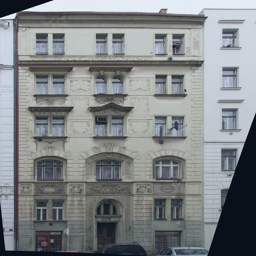} &\
    \includegraphics[width=0.16\textwidth]{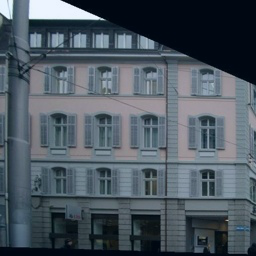} &\
    \includegraphics[width=0.16\textwidth]{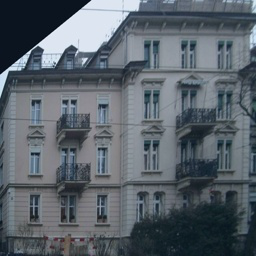} \\

    \multicolumn{5}{l}{ Wasserstein Reproduction (Baseline):} \\
    \includegraphics[width=0.16\textwidth]{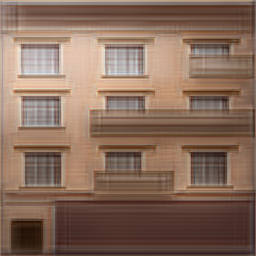} &\
    \includegraphics[width=0.16\textwidth]{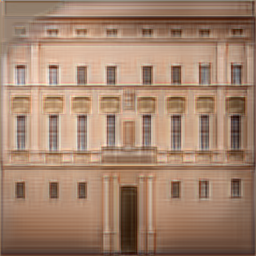} &\
    \includegraphics[width=0.16\textwidth]{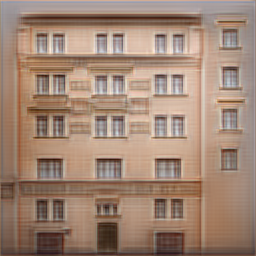} &\
    \includegraphics[width=0.16\textwidth]{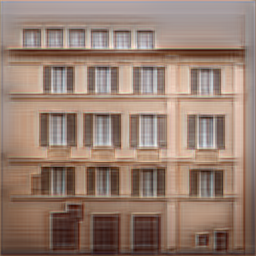} &\
    \includegraphics[width=0.16\textwidth]{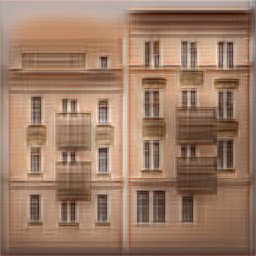} \\

    \multicolumn{5}{l}{ TaylorGAN Reproduction:} \\
    \includegraphics[width=0.16\textwidth]{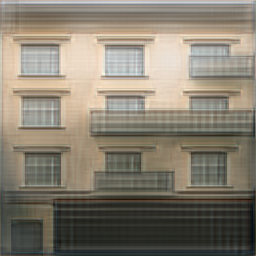} &\
    \includegraphics[width=0.16\textwidth]{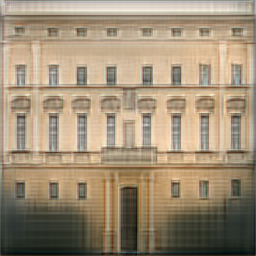} &\
    \includegraphics[width=0.16\textwidth]{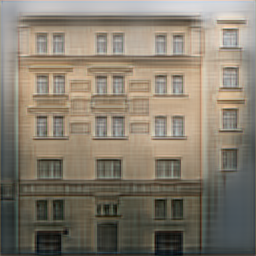} &\
    \includegraphics[width=0.16\textwidth]{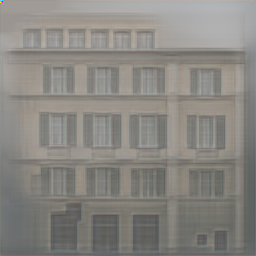} &\
    \includegraphics[width=0.16\textwidth]{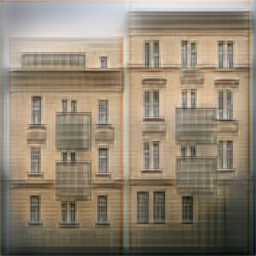} \\

  \end{tabular}
  \caption{\textbf{Five random samples from the CMP Facade test dataset, comparing Wasserstein and TaylorGAN loss functions.} The loss functions are used to train pix2pix-HD models that take architectural element annotations (top row) and generate corresponding photorealistic images similar to the ground-truth (second row). Images from the model trained with TaylorGAN (bottom row) have a higher quality than the baseline (third row). TaylorGAN images have more realistic coloration, better separation of the buildings from the sky, and finer details than the baseline.}
  \label{fig:gloganresults}
\end{figure*}

TaylorGAN found a set of loss functions that outperformed the original Wasserstein loss with spectral normalization. After 49 generations of evolution, it discovered the loss functions
\begin{equation}
    \begin{aligned}
\mathcal{L}_{D_{\text{real}}} =  5.6484\;(\Dx - 8.3399) + 9.4935\;(\Dx - 8.3399)^2 \\+ 8.2695\;(\Dx - 8.3399)^3
    \end{aligned}
\end{equation}
\begin{equation}
    \begin{aligned}
\mathcal{L}_{D_{\text{fake}}} =  6.7549\;(\DGz - 8.6177) + 2.4328\;(\DGz - 8.6177)^2 \\+ 8.0006\;(\DGz - 8.6177)^3
    \end{aligned}
\end{equation}
\begin{equation}
    \begin{aligned}
\mathcal{L}_{G_{\text{fake}}} =  0.0000\;(\DGz - 5.2232) + 5.2849\;(\DGz - 5.2232)^2 \\+ 0.0000\;(\DGz - 5.2232)^3 .
    \end{aligned}
\end{equation}
A learning rate of $0.0001$, discriminator feature loss weight of $4.0877$, and perceptual distance loss weight of $10.3155$ evolved for this candidate.

Figure~\ref{fig:gloganresults} compares images for five random test samples that were generated with both the Wasserstein baseline and metalearned TaylorGAN loss functions. Visually, the TaylorGAN samples have more realistic coloration and details than the baseline. Baseline images all have an orange tint, while TaylorGAN images more closely match ground-truth images' typical coloration. Note that color information is not included in the model's input, so per-sample color matching is not possible. Additionally, TaylorGAN images tend to have higher-quality fine-grained details. For example, facade textures are unnaturally smooth and clean in the baseline, almost appearing to be made of plastic.

Quantitatively, the TaylorGAN model also outperforms the Wasserstein baseline. Across ten Wasserstein baseline runs, the average test-set SSIM was $9.4359$ and the average test-set perceptual distance was $2129.5069$. The TaylorGAN model improved both metrics, with a SSIM of $11.6615$ and perceptual distance of $2040.2561$. 

Notably, the training set is very small, with fewer than 500 image pairs, showing how loss-function metalearning's benefits on small classification datasets also extend to GANs. Thus, metalearned loss functions are an effective way to train better GAN models, extending the types of problems to which evolutionary loss-function metalearning can be applied.

\section{Discussion and future work}
\label{sec:discussion}

The results in this paper show that evolving GAN formulations is a promising direction for research. On the CMP Facade benchmark dataset, TaylorGAN discovered powerful loss functions: With them, GANs generated images that were qualitatively and quantitatively better than those produced by GANs with a Wasserstein loss. This unique application showcases the power and flexibility of evolutionary loss-function metalearning, and suggests that it may provide a crucial ingredient in making GANs more reliable and scalable to harder problems.

At first glance, optimizing GAN loss functions is difficult because it is difficult to quantify a GAN's performance. That is, performance can be improved on an individual metric without increasing the quality of generated images. Multiobjective evolution, via composite objectives, is thus a key technique that allows evolution to work on GAN formulations. That is, by optimizing against multiple metrics, each with their own negative biases, the effects of each individual metric's bias will not deleteriously affect the path evolution takes.

There are several avenues of future work with TaylorGAN. First, it can naturally be applied to different datasets and different types of GANs. While image-to-image translation is an important GAN domain, there are many others that can benefit from optimization, such as image super-resolution and unconditioned image generation. Since TaylorGAN customizes loss functions for a given task, dataset, and architecture, unique sets of loss functions could be discovered for each of them.

There is a wide space of metrics, such as Delta E perceptual color distance \citep{robertson1990historical}, that quantify different aspects of image quality. They can be used to evaluate GANs in more detail and thus guide multiobjective evolution more precisely, potentially resulting in more effective and creative solutions.

\section{Conclusion}
\label{sec:conclusion}

While GANs provide fascinating opportunities for generating realistic content, they are difficult to train and evaluate. This paper proposes an evolutionary metalearning technique, TaylorGAN, to optimize a crucial part of their design automatically. By evolving loss-functions customized to the task, dataset, and architecture, GANs can be more stable and generate qualitatively and quantitatively better results. TaylorGAN may therefore serve as a crucial stepping stone towards scaling up GANs to a wider variety and harder set of problems.

\bibliographystyle{abbrvnat}
\bibliography{bib-prop,bib-gan,bib-software}


\end{document}